\documentclass[letterpaper, 10 pt, conference]{ieeeconf}  

\IEEEoverridecommandlockouts                              

\usepackage{graphics} 

\usepackage{graphicx}
\usepackage{subfig}
\usepackage{epstopdf}
\usepackage{float}

\title{\LARGE \bf
A bistable soft gripper with mechanically embedded sensing and actuation for fast closed-loop grasping 
}

\author{Thomas George Thuruthel$^{1}$, Syed Haider Abidi$^{1}$, Matteo Cianchetti$^{1}$, Cecilia Laschi$^{1}$ and Egidio Falotico$^{1}$
\thanks{$^{1}$BioRobotics Institute, Scuola Superiore Sant'Anna, Viale Rinaldo Piaggio 34, Pontedera (Pisa), Italy}%
}

\begin{document}

\maketitle
\thispagestyle{empty}
\pagestyle{empty}

\begin{abstract}
Soft robotic grippers are shown to be high effective for grasping unstructured objects with simple sensing and control strategies. However, they are still limited by their speed, sensing capabilities and actuation mechanism. Hence, their usage have been restricted in highly dynamic grasping tasks. This paper presents a soft robotic gripper with tunable bistable properties for sensor-less dynamic grasping. The bistable mechanism allows us to store arbitrarily large strain energy in the soft system which is then released upon contact. The mechanism also provides flexibility on the type of actuation mechanism as the grasping and sensing phase is completely passive. Theoretical background behind the mechanism is presented with finite element analysis to provide insights into design parameters. Finally, we experimentally demonstrate sensor-less dynamic grasping of an unknown object within 0.02 seconds, including the time to sense and actuate.  
\end{abstract}

\section{Introduction}

Soft robotic technologies have been very effective in grasping tasks \cite{shimoga1996soft,kim2013soft,shintake2018soft}. This is because of their conformability to uneven and unstructured objects and their ability to dissipate impulse forces generated during a grasp \cite{shimoga1996soft}. This allowed to develop simple, mechanically adaptive systems that excel at gripping complex objects as shown in \cite{brown2010universal}. The field of soft robotic grippers has lately highly diverged with emphasis on actuation technologies \cite{chen2017electronic,shintake2015variable}, material properties \cite{taccola2015toward,shan2013soft,martinez2014soft,shepherd2013soft}, fabrication techniques \cite{yap2016high}, optimal design and analysis \cite{liu2018optimal,zhou2015soft,deimel2016novel}, gripping mechanisms \cite{song2014soft,glick2018soft}, sensing technologies \cite{truby2018soft,bilodeau2015monolithic,shih2017custom}, modelling \cite{Thurutheleaav1488,soter2018bodily} and innovative applications \cite{galloway2016soft,wang2017prestressed,do2018soft}. Some of the current challenges in soft robotic grippers include speed, integrated sensing and actuation mechanism \cite{shintake2018soft}. This paper provides simple solution to the aforementioned challenges in a purely mechanical way operating on the concept of morphological computation \cite{pfeifer2009morphological}.

The actuation bandwidth of a soft gripper is determined by the gripper material and the actuation mechanism. The response time of the gripper can be reduced by effective force transmission from the actuation mechanism. The fastest reported actuation mechanism found in literature was a simple finger with electrorheological fluid sandwiched between a grounded elastomeric skin and a charged conducting cathode \cite{kenaley1989electrorheological}. As the fluid reacts rapidly to the high voltage source leading to a response time of 0.001 seconds. Other works with comparable response time involving differential pressure \cite{krahn2017soft}, dielectric elastomeric actuator \cite{lau2017dielectric}, granular jamming \cite{amend2016soft} and controlled adhesion \cite{hawkes2015grasping,shintake2016versatile} could only reach a response time of 0.1 seconds. None the less, it must be noted that the minimization of the gripper response time was not the main objective of these works. This works presents a passive grasping mechanism. Hence, the speed of actuation is solely determined by the material properties and the effective impedance of the gripper along the direction of actuation.  

Majority of soft robotic grippers are controlled in open-loop without any feedback before and after contact with the object. In these cases, they rely on certain user-provided information on the location and strategy for successful grasp. Embedded sensing can automate the process even more, by providing additional information about the location, type and physical properties of the object \cite{wang2018toward}. Embedded sensing for pre-touch information can be used for deciding the closing pattern \cite{tavakoli2017autonomous}, for detection and identification of grasped objects \cite{ho2017design,homberg2015haptic}, and closed-loop control of the soft finger \cite{luo2017toward}. However, the response rate of these sensors are generally in the order of 100 ms \cite{shintake2018soft}, limited by the slow response of the soft mechanical system and data processing loop. Moreover, obtaining accurate distributed sensing capabilities is still a hard challenge in soft robotics. In this paper, we introduce a purely mechanical feedback system for the closed-loop control of the soft gripper. Arguably, this would be the fastest possible way to close the loop using contact information. 

This work introduces a soft robotics gripper with tunable bistable properties. The bistable mechanism allows us to store arbitrarily large strain energy in the soft system which can then be easily released upon contact. Using this simple mechanism we are able to demonstrate a sensor-less, passive, closed-loop grasping of unknown objects within 0.02 seconds. The mechanism also provides flexibility on the type of actuation mechanism as the grasping and sensing phase is completely passive. Hence, we can provide arbitrarily high grasping forces while preserving the natural stiffness of the gripper material. Moreover, since the open and closed state of the gripper is a stable equilibrium point, no power is consumed while the gripper holds an object or when the gripper is open. 
\begin{figure*}[t]
\centering
  \includegraphics[width=0.8\linewidth]{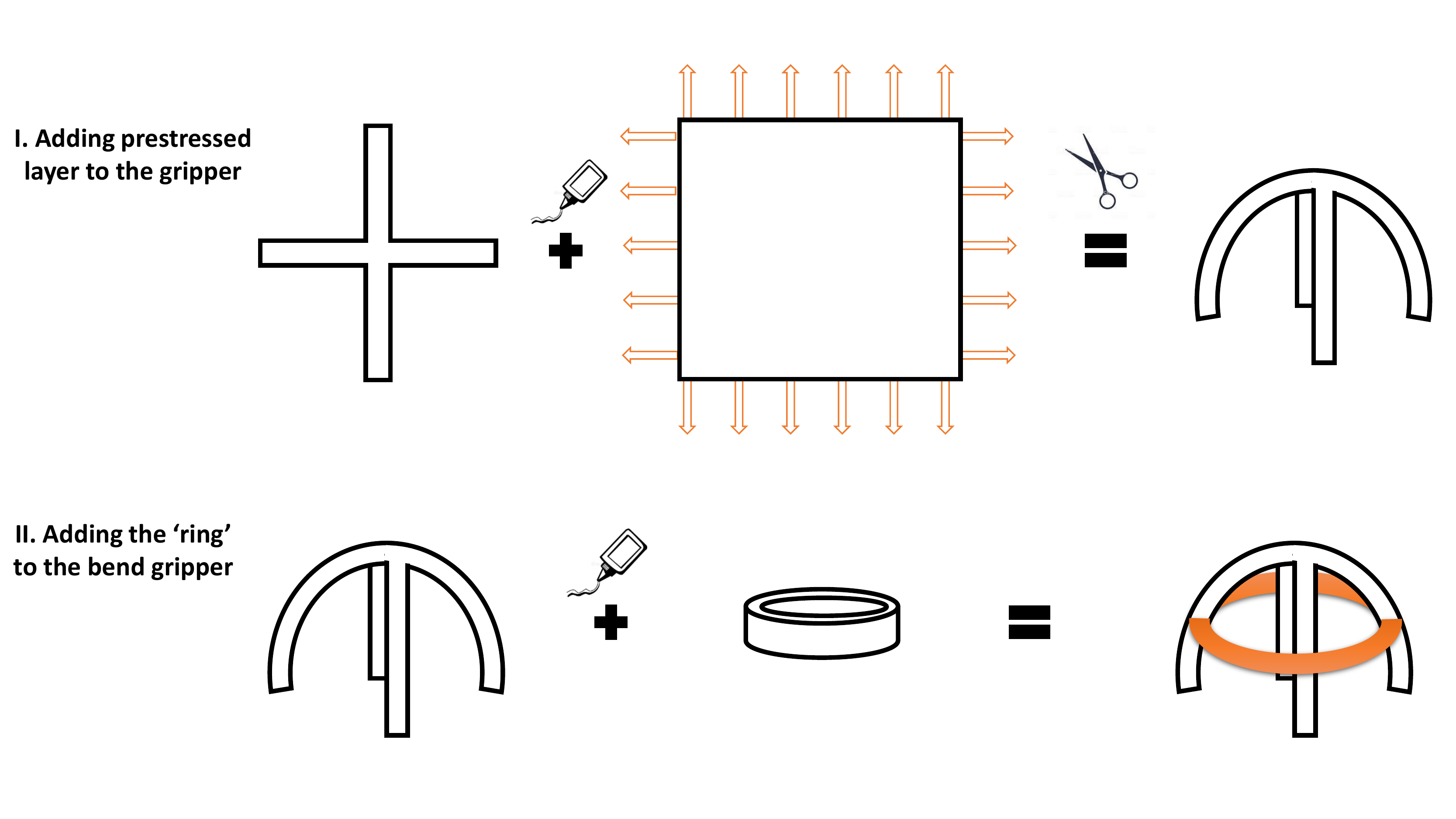}
  \caption{Fabrication process of the bistable gripper. }
  \label{fig:making}
\end{figure*}

\subsection{Bistable Mechanisms in Robotics}

A bistable mechanism is a system with two stable equilibrium points. In a conservative system this corresponds to two local minima in the total potential energy of the system. This property is commonly used in the design of mechanical switches, flip-flops, etc. In robotics, bistability is often used for reducing control complexity, for performing fast motions, and for energy conservation. The impulse forces generated when the bistable mechanism transitions through the snap-through region has been used for propulsion \cite{noh2012flea,chen2018harnessing}. Recently, a soft bistable vale mechanism was described for soft actuators \cite{rothemund2018soft}. Dielectric actuators are often used with bistable mechanisms for light weight, large displacement applications \cite{wingert2006design,follador2015design}. A number of designs for bistable gripping has also been proposed over the years \cite{wang2016soft,kim2010towards,follador2015bistable}. However, these mechanisms are based on bistablity of their individual fingers and hence do not provide the high conformability required from a soft gripper. Additionally, their snap-through energy cannot be tuned desirably which makes embedded mechanical sensing challenging. We present a design methodology for creating fully soft bistable grippers. The simple mechanism preserves all the conformability of a soft gripper and provides new paradigms for field of soft sensing.

\section{Manufacturing}
The fabrication of the bistable gripper begins with molding of the base soft gripper (Figure \ref{fig:making}). The mold for the base gripper was 3D printed. The base gripper was made from a silicone elastomer (Dragon Skin 30, Smooth-On Inc.). Then the chamber inside the gripper was sealed with a thin bottom layer of the same material. Simultaneously, a thin layer of another silicone elastomer (Ecoflex 30, Smooth-On Inc.) was cured, stretched in all directions and clamped. The base gripper was then attached to the strained layer using a silicone adhesive (Sil-Poxy, Smooth-On Inc.) (Figure \ref{fig:making}). The tube for pneumatic actuation was also added during this period. The pneumatic chamber is only kept for fine tuning the two equilibrium states of the final mechanism which will be described later. For this work, we do not use any active actuators.  

Once the pre-stressed layer was bonded, the clamp on the stretched layer was removed. The causes the gripper to bend uniformly and reach its new equilibrium position. A thin layer of silicone elastomer (Ecoflex 30, Smooth-On Inc.) was then cut into strips, rolled around and bonded to obtain the ring. This ring was wrapped on the gripper and bonded using the silicone adhesive Figure \ref{fig:making}). To destabilize the open state of the bistable mechanism, the ring can be trimmed, which reduces its stored elastic energy. To increase the gripping force, additional rings can be added similarly.
\begin{figure*} [t]
\centering
  \includegraphics[width=\linewidth]{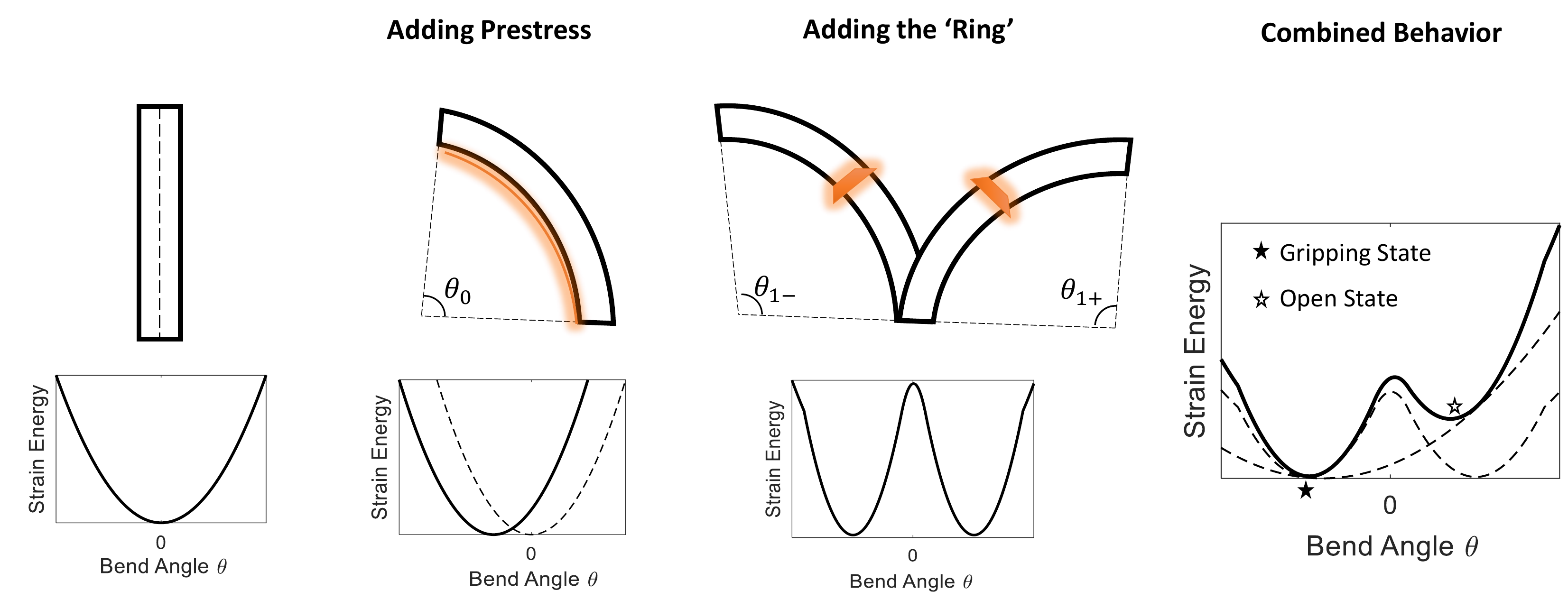}
  \caption{Elementary model of the gripper working principle.}
  \label{fig:theory}
\end{figure*}


\section{Theory}

The working principle of the proposed bistable gripper is described in this section. Bistability arises when a system has two stable equilibrium states. These states would correspond to two local minima of the potential energy. Hence, our objective is to develop a soft gripper with two stable equilibrium states. The first state should correspond to the gripping configuration with the lowest stored potential energy (For higher forces) and a stable open configuration with a small snap-through energy. This allows the gripper to transition from the open state to the closed state with very low injection of energy into the system. To understand the design of our proposed mechanism, we first model the mechanism as two independent systems and combine them using the principle of superposition (See figure \ref{fig:theory}). The fingers of our mechanism can be modelled as elastic beams and the ringed system can be modelled as thin elastic bands. All the analysis is presented in two dimensions for easy visualization.   

Ignoring the contribution of gravitational energy the steady state potential energy of a beam with bend angle $\theta$ would be the stored elastic strain energy :

\begin{equation}
    U=\frac{ M\theta}{2} \\
    = \frac{ \theta^2EI}{2L} \propto\theta^2
\end{equation}

Where $M,E,I,L$ is the applied moment, Youngs Modulus, moment of inertia and the length of the beam. Adding a uniformly pre-stressed element to the beam shifts the equilibrium configuration of the beam (Figure \ref{fig:theory}). This shift in equilibrium position can also be achieved through the manufacturing process with appropriate molds. Our manufacturing process reduces the complexity and size of the finger mold, but adds an additional step in the pre-straining process. 

\begin{equation}
    U\propto {(\theta-\theta_0)}^2
\end{equation}

To increase the force applied by the gripper, we can increase the pre-stress added to the system. If higher forces are required with the same bend angle, the Youngs Modulus of the base material can be increased accordingly. 

Considering the ringed system independently by ignoring the elastic contribution of the bent beam and considering only its kinematic contribution, the ring has two low energy states. This occurs at the two mirror states in which the ring has its initial radius. Hence, in a way, the ring mechanism is bistable by itself and would have two stable bend angle configurations $\theta_{1-},\theta_{1+}$ (Figure \ref{fig:theory}). Combining these two systems, using the principle of superposition, provides us with a bistable mechanism with our required behavior. Tuning of the strain energy curves can be achieved by playing with the parameters of the pre-stress mechanism, the ring mechanism and the material properties. We summarize them here and using finite element analysis.

If $\theta_{1-}$ is higher in magnitude than $\theta_{0}$, the applied forces will increase accordingly with the corresponding shift in the equilibrium position. By increasing the magnitude of $\theta_{0}$ or by reducing the magnitude of $\theta_{1-}$ or by reducing the Youngs Modulus of the ring material, the stability of the \textit{open} state can be reduced. The same can be achieved by reducing the thickness/width of the ring. In other words, we have to reduce the stored elastic energy in the isolated ring mechanism. The morphology and non-linear material properties of the gripper also plays a role in the bending profile, speed and force. Numerical validation of our hypotheses is shown in the next section.


\section{Finite Element Analysis}

Finite element analysis (FEA) of a geometrically simplified model of the gripper is performed to study the effect of different design parameters. Primarily we study the effect of the finger curvature (or pre-strain), the ring geometry and placement on the gripping force, snap-through energy and the time to close. Our interests lie on the relative potential energy values of the stable open-state and the unstable transition peak with respect to the base potential energy value at the stable closed state (Figure \ref{fig:theory})

\begin{figure} [h]
\centering
  \includegraphics[width=0.8\linewidth]{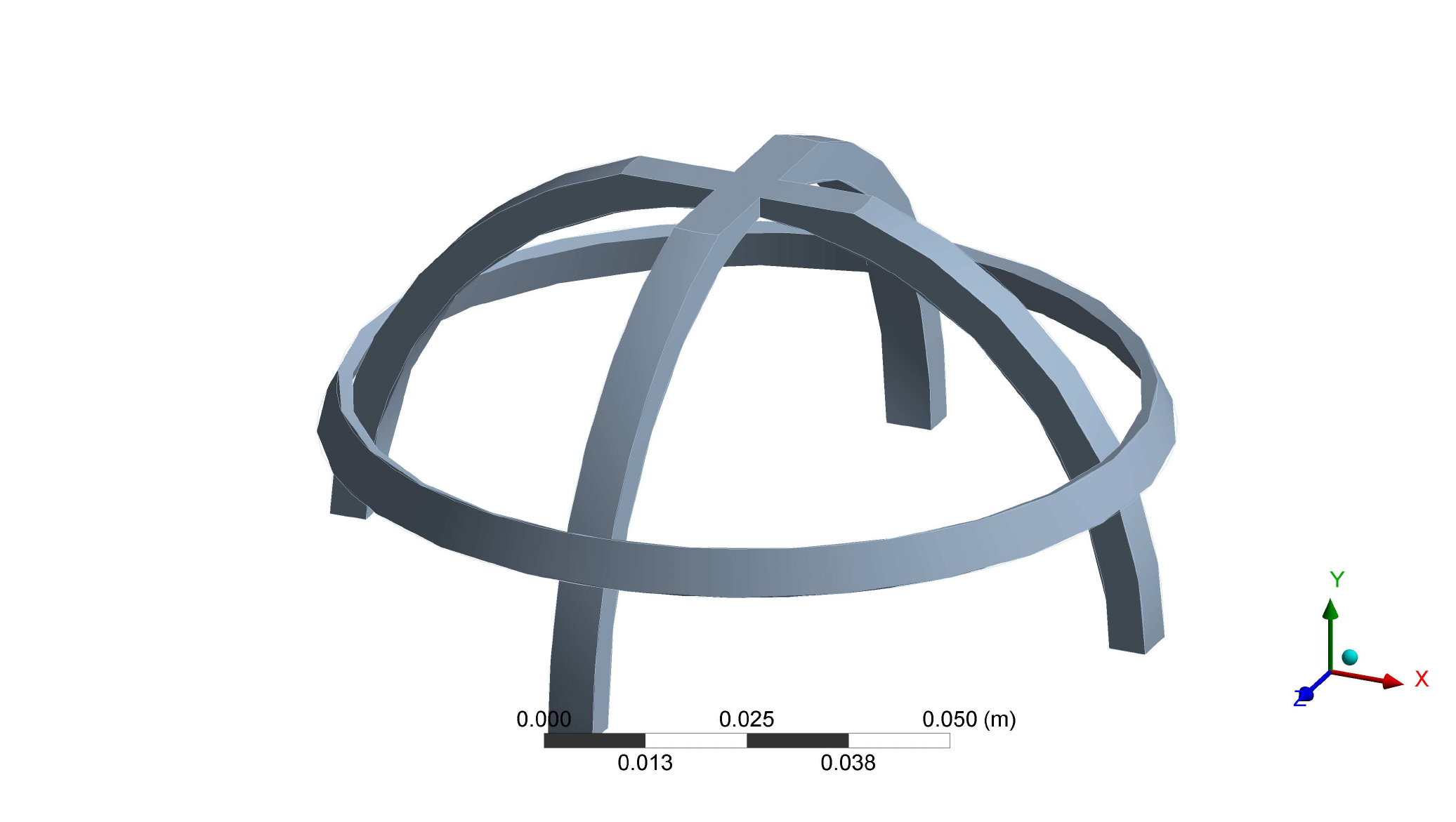}
  \caption{Geometry of the gripper used for the FEA.}
  \label{fig:fea}
\end{figure}

The simplified model of the gripper used for the FEA is shown in figure \ref{fig:fea}. All the FEA is performed on \textit{ANSYS}. The elastomer sample (Yeoh) material is used for the whole structure. The pre-strain energy obtained from our manufacturing process is not modelled for simplicity. If the material exhibited linear elastic properties, this assumption does not affect the relative potential energy values. In other words, both the pre-strained bent finger and a molded bent finger would be equivalent in performance. However, due to the nonlinear properties of an elastic material, there will be differences among the two. We do not model the pneumatic chambers for our analysis too. 

Four cases of morphological changes were studied with respect to the original morphology. The cases were : placing the ring higher (towards the base), placing the ring lower (towards the tip), reducing the thickness of the ring, and increasing the curvature of the fingers (from 0.20$cm^{-1}$ to 0.25$cm^{-1}$). The point of this study is to develop design criterion's for shaping the gripper properties.

\begin{figure} [h]
\centering
  \includegraphics[width=0.8\linewidth]{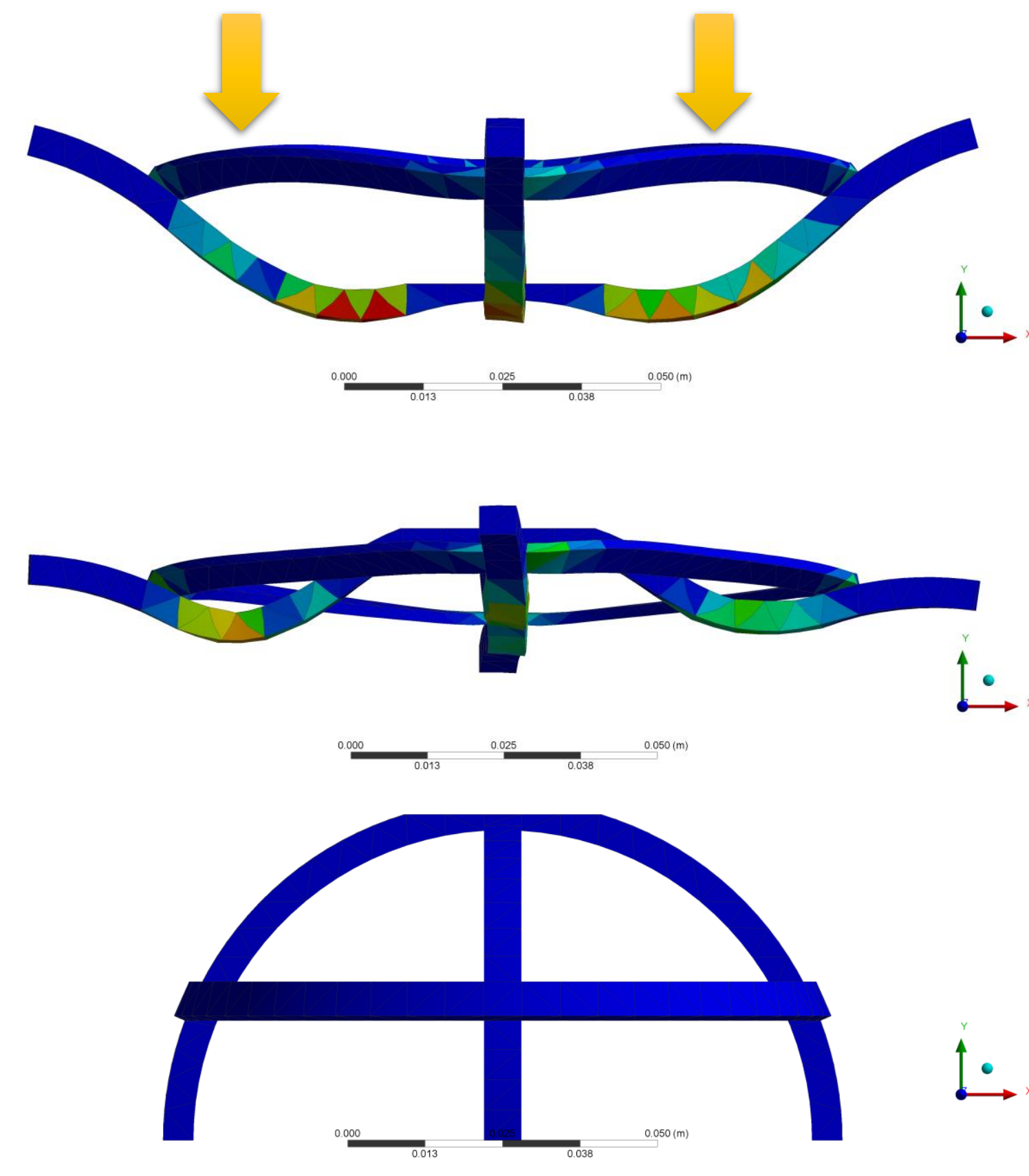}
  \caption{Loading constraint applied on the finite element model to study the static strain energy dependencies on the design parameter.}
  \label{fig:constraints}
\end{figure}

Static analysis of the model was performed and the total strain energy in the structure was calculated for each cases. First, all the models were provided multi-dimensional displacement constraints that brought the gripper to the open stable state (Figure \ref{fig:constraints}). Then, one-dimensional ramped forces were applied to the ring as shown in figure \ref{fig:constraints} until the gripper reached the closed stable state again. Stabilization is turned on near the unstable equilibrium point for better convergence and it is ensured that the total stabilization energy is significantly less than the total strain energies for validity of the results. 

\begin{figure} [h]
\centering
  \includegraphics[width=\linewidth]{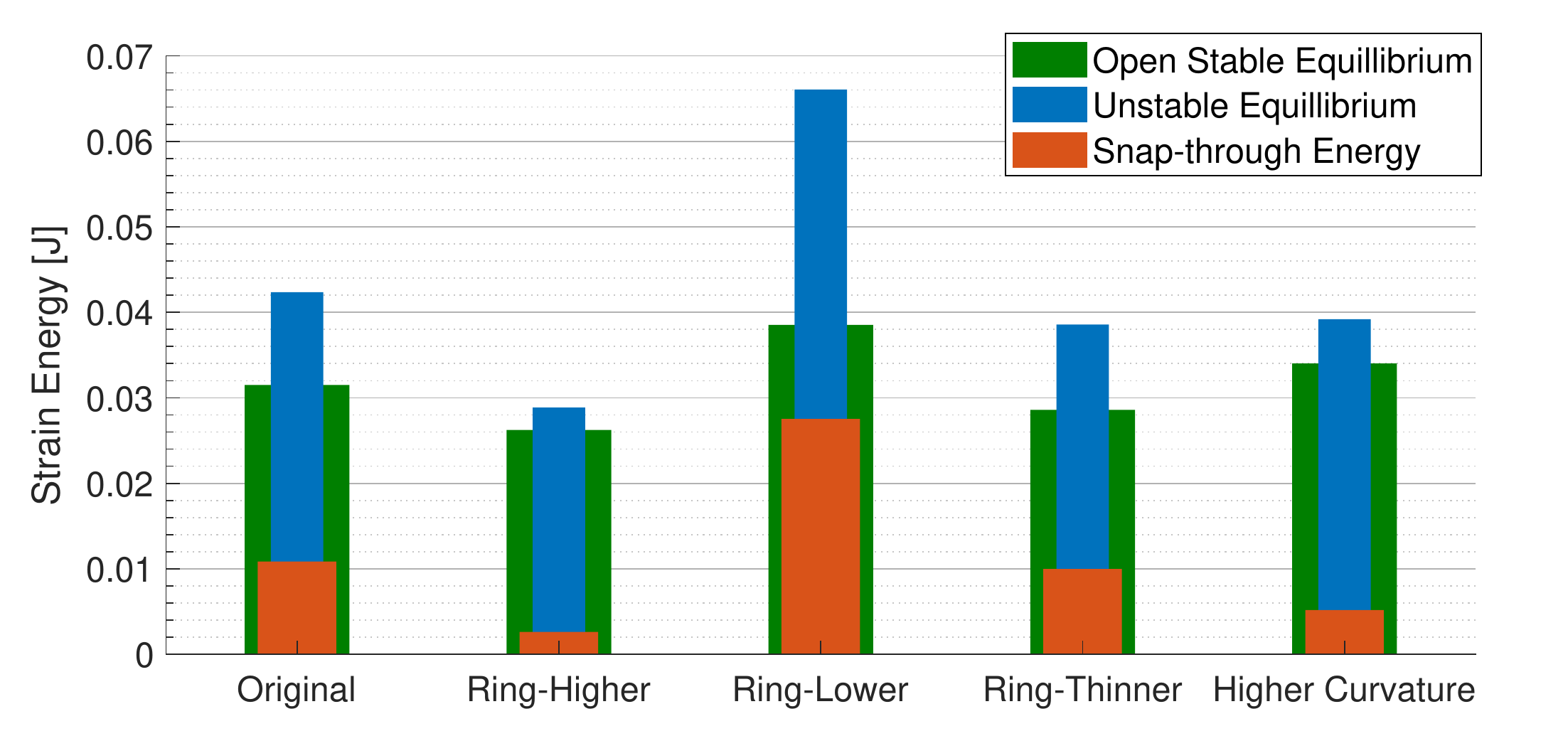}
  \caption{Strain energy estimates obtained from the FEA for the design parameters of concern. Snap-through energy is calculated as the difference between the unstable state strain energy and the open stable state strain energy.}
  \label{fig:feares}
\end{figure}

The total strain energy values for each design cases are shown in Figure \ref{fig:feares}. Placing the ring higher, as expected, reduces the stored elastic energy at the open state and hence reduces the snap-through energy. This is because the radius of the ring reduces as it is displaced higher and hence the opposing forces that the ring provides at the open state reduces. At the closed state, the curvature of the finger and stored strain energy remains the same, however, the effective stiffness of the gripper will reduce and hence would apply lesser gripping forces with respect to the original morphology. Placing the ring lower, conversely, increases the strain energy at the open state and the snap-through energy and the gripping forces. Assuming all the three designs transition around the same configuration, the time for the gripper to close should be inversely proportional to the equivalent stiffness of the mechanism (As the motion is completely passive, the time to close should be inversely proportional to the natural frequency of the mechanism). Therefore, the mechanism with lower ring placement would close the fastest.

Reducing the ring thickness (by half) reduces the stored potential energy at the open state and the snap-through energy, however, not so significantly. Therefore, the ring thickness or width could be final parameter to be adjusted to fine tuned for higher precision (As demonstrated in the experimental section). Increasing the curvature of the beam increases the potential energy of the open state, but reduces the snap-through energy. For our manufacturing technique this can be achieved by increasing the pre-strain amount. As the curvature of the gripper increases, the gripping forces would also increase as the gripper would be displaced by a higher amount, for a fixed object size, when compared to the original morphology. So by increasing the beam curvature and accordingly displacing the ring lower, we can arbitrarily increase the gripping force while maintaining the same snap-through energy. This would result in a decrease in the closing time also. It must be noted that we are ignoring all the dynamic effects like damping and momentum which would also affect the closing speed.   

   \begin{figure*} [h]
  \includegraphics[width=\linewidth]{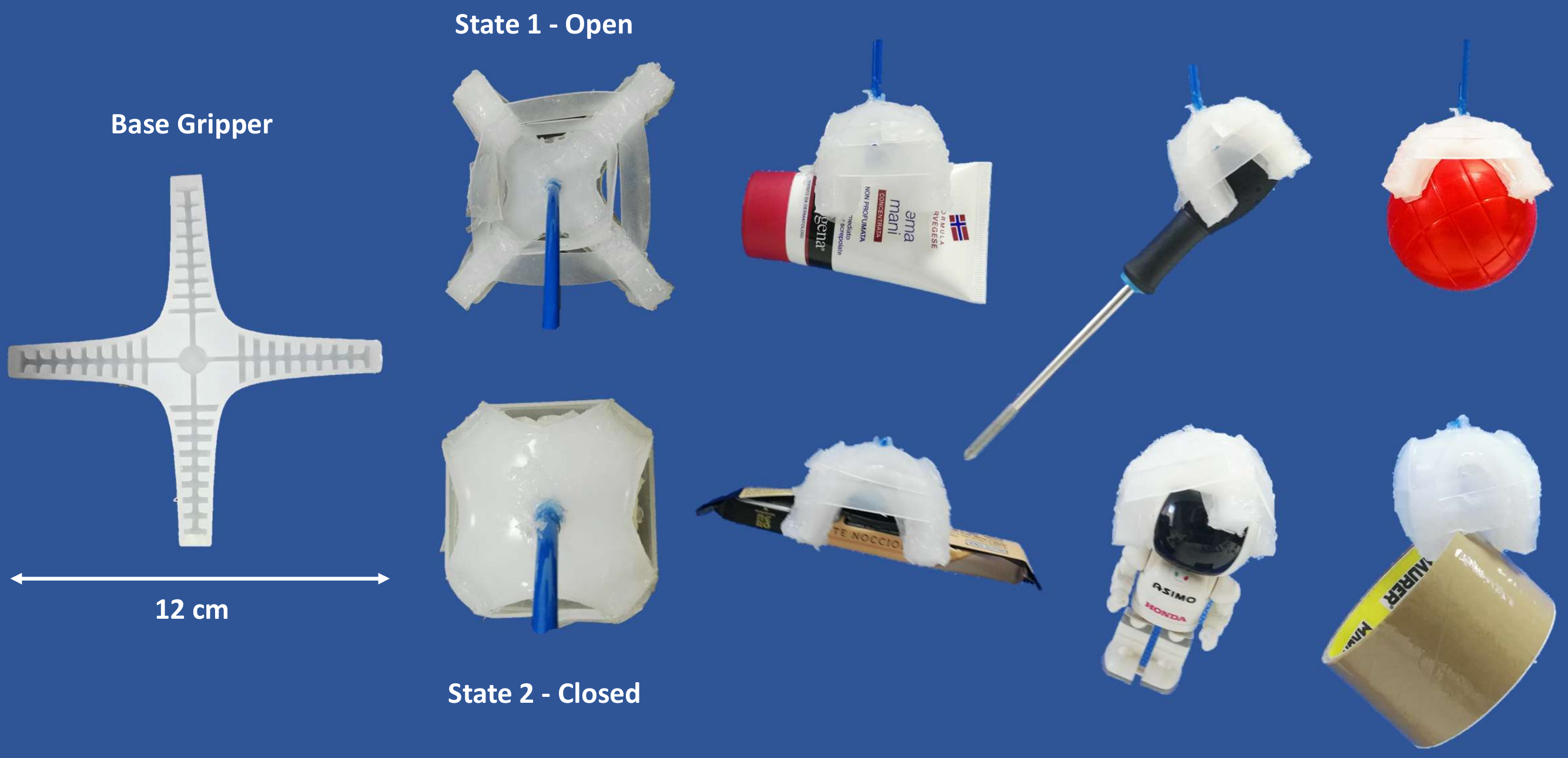}
  \caption{Gripper prototype and demonstration of the gripper's ability to grasp objects of different shape and size}
  \label{fig:demo}
\end{figure*}

\section{Experimental Results}

Two sets of experiments are conducted to validate our concept of the bistable soft gripper. The first one is the task of autonomously grasping a dynamic unstructured object using the concept of mechanical embedded sensing. The second set of tests are performed to demonstrate the soft grippers ability to grasp static objects of varying shapes and hold them without further energy requirements. For all the experiments, the gripper is manually reset to its open stable state. The actuation is completely passive using only the stored elastic energy for motion.   

\subsection{Embedded sensing and grasping}

\begin{figure} [h]
   \centering
  \includegraphics[width=\linewidth]{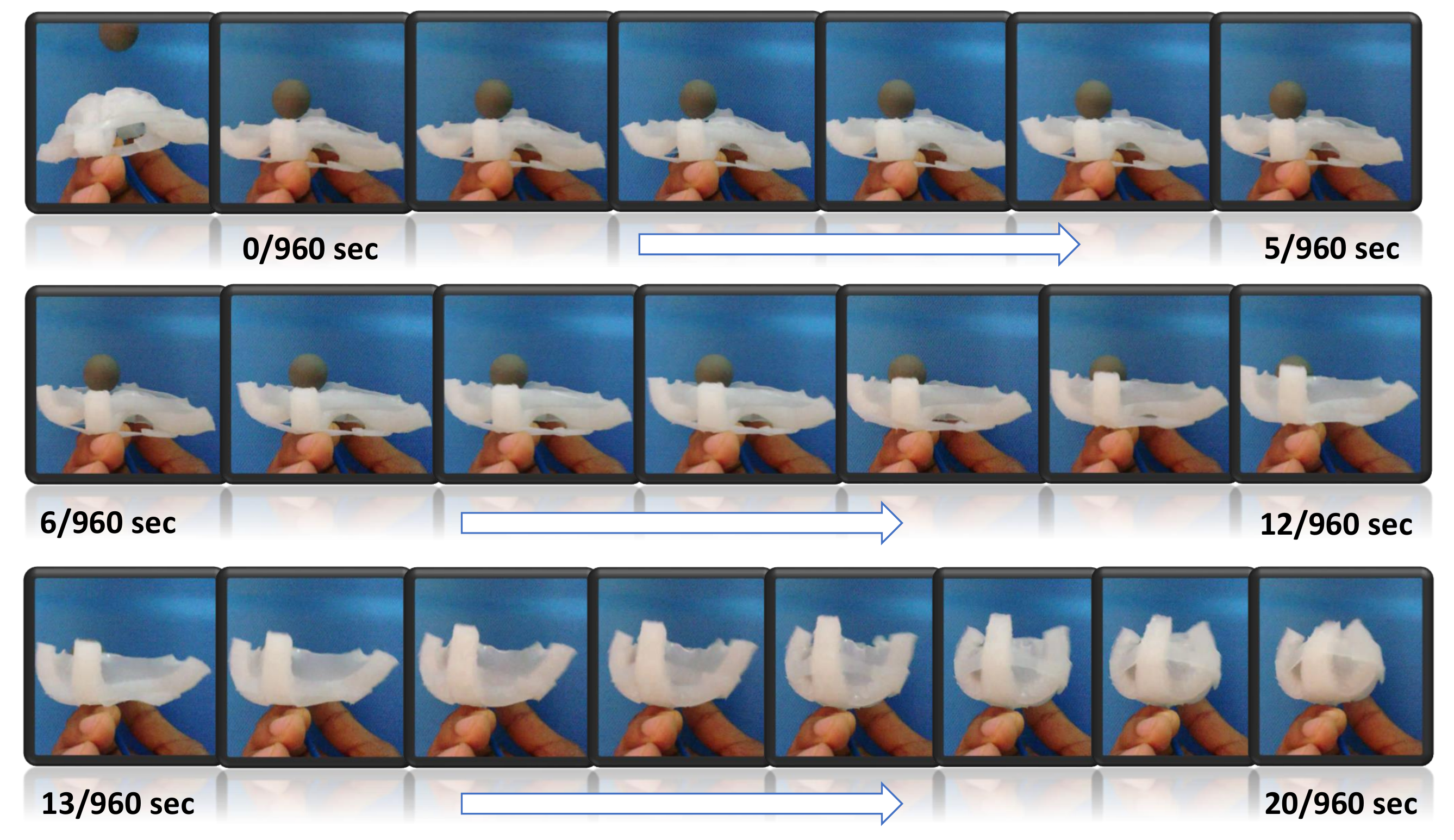}
  \caption{Autonomous object detection and closing}
  \label{fig:ball}
\end{figure}

The transition of the gripper from the open stable state to the closed stable state happens by applying a minimum required force to the system in the right direction. The direction of the force should be in the direction as shown in figure \ref{fig:constraints}. For our tests, we obtain this from the reaction forces acting on the gripper when a object comes in contact with the face of the gripper (see figure \ref{fig:ball}). The results are analyzed visually using a 960 frames per second video footage. The time elapsed from physical contact of the object and the complete closure of the gripper is 0.021 seconds (See supplementary video). 

\begin{figure} [h]
  \includegraphics[width=\linewidth]{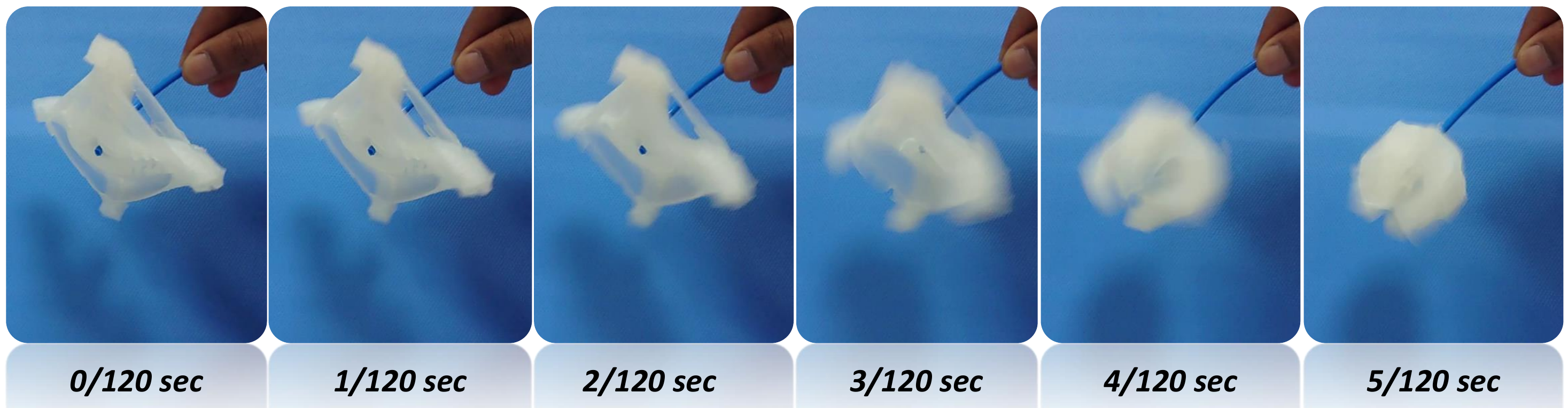}
  \caption{Tuning the snap-through energy so that the gripper closes just by gravitational forces}
  \label{fig:grav}
\end{figure}

To demonstrate the ease in tuning the gripper properties through physical modifications, we modify the original gripper to have very low snap-through energy. We do this by manually clipping the ring width, so that the stored strain energy in the open stable state reduces as shown in figure \ref{fig:feares}. We reduce the width until the gripper activates just by gravitational forces (See figure \ref{fig:grav}). As the equivalent stiffness of the gripper reduces by trimming the ring width, the time to close increases, as predicted. So for this case, the time to close is between 0.03-0.04 seconds. 

By modifying the geometry of the gripper and the manipulator to which it is attached to, other sensor-less grasping strategies can be investigated. Conversely, the bistable properties of the gripper can be easily inverted to make the open state more stable. Hence, by tuning the snap-through energy opening of the gripper can be autonomously achieved by just controlling the orientation of the gripper base.  

\subsection{Grasping objects}

As the gripper is made solely using a soft elastomeric material and the ring structure lies outside the gripping surface, all the advantages of a completely soft gripper is applicable to our design. The gripping forces applied by the arm can be tuned using the methods prescribed using the FEA. The minimum size of the object grasped is dependent on the pre-strain in the finger and the pre-strain (if any) in the ring. In short, it depends on the final curvature (shape) of the finger in the closed state. Figure \ref{fig:demo} shows the various objects the gripper can hold. Note that no energy is consumed once the gripper reaches the closed-state and can therefore can hold the object indefinitely.

\section{Conclusion}

This paper presents the design and development of a bistable soft gripper for sensor-less dynamic grasping. The mechanism is an operative demonstration of the concept of \textit{morphological computation} or \textit{embodied intelligence}. This is shown for the task of closed-loop control of a fully soft gripper. All the sensing and control required for closed-loop control is mechanically embedded. Hence, for the given material properties, we can achieve one of the fastest possible closed-loop reactive grasping strategy. Other closed-loop strategy will be slowed down because of the delays involved in transferring and processing the sensor signals added with the delays involved in actuating the actuation mechanism. Predictive grasping strategies, on the other hand, could provide faster responses, however, they involve more complexities. Due to the passive nature of closing actuation, the natural compliance of the gripper material and structure is preserved during a grasp \cite{della2017controlling}. As the mechanism requires external power only for opening the gripper, the bistability in the mechanism is highly desirable for energy efficiency. No power is used for sensing, transitioning, grasping and for staying open. As there is no energy conversion/transfer from the actuator to the soft mechanism (it is elastically stored in the elastomer), the release of energy during closure is also highly energy efficient. 

We present design heuristics using simplified models for the proposed soft gripper that allows the user to tune basic gripper properties like snap-through energy, grasping forces and closing time. However, our analysis is based on static forces assuming passive motion dynamics. Dynamic properties like inertia and damping would also affect the passive grasping characteristics. Another area of concern is the increased induced creep in the elastomeric material due to the pre-strain in the mechanism. This can be avoided by an appropriate starting mold (as shown in the FEA), nonetheless, it restricts the margin for tuning the gripper properties. 

Future works include alternate designs for the gripper that have more complex passive closing behaviors aimed towards autonomous grasping as well as manipulation. This would require numerical or analytical dynamical studies on different finger and ring geometries. The gripper is brought to the open stable state manually in this work. Actuation mechanisms that can perform this task must be investigated for automating this process. Tendon driven mechanism that run along the four arms is a potential solution.



\begin{thebibliography}{10}

\bibitem{amend2016soft}
John Amend, Nadia Cheng, Sami Fakhouri, and Bill Culley.
\newblock Soft robotics commercialization: Jamming grippers from research to
  product.
\newblock {\em Soft robotics}, 3(4):213--222, 2016.

\bibitem{bilodeau2015monolithic}
R~Adam Bilodeau, Edward~L White, and Rebecca~K Kramer.
\newblock Monolithic fabrication of sensors and actuators in a soft robotic
  gripper.
\newblock In {\em Intelligent Robots and Systems (IROS), 2015 IEEE/RSJ
  International Conference on}, pages 2324--2329. IEEE, 2015.

\bibitem{brown2010universal}
Eric Brown, Nicholas Rodenberg, John Amend, Annan Mozeika, Erik Steltz,
  Mitchell~R Zakin, Hod Lipson, and Heinrich~M Jaeger.
\newblock Universal robotic gripper based on the jamming of granular material.
\newblock {\em Proceedings of the National Academy of Sciences},
  107(44):18809--18814, 2010.

\bibitem{chen2017electronic}
Dustin Chen and Qibing Pei.
\newblock Electronic muscles and skins: A review of soft sensors and actuators.
\newblock {\em Chemical reviews}, 117(17):11239--11268, 2017.

\bibitem{chen2018harnessing}
Tian Chen, Osama~R Bilal, Kristina Shea, and Chiara Daraio.
\newblock Harnessing bistability for directional propulsion of soft, untethered
  robots.
\newblock {\em Proceedings of the National Academy of Sciences}, page
  201800386, 2018.

\bibitem{deimel2016novel}
Raphael Deimel and Oliver Brock.
\newblock A novel type of compliant and underactuated robotic hand for
  dexterous grasping.
\newblock {\em The International Journal of Robotics Research},
  35(1-3):161--185, 2016.

\bibitem{della2017controlling}
Cosimo Della~Santina, Matteo Bianchi, Giorgio Grioli, Franco Angelini, Manuel
  Catalano, Manolo Garabini, and Antonio Bicchi.
\newblock Controlling soft robots: balancing feedback and feedforward elements.
\newblock {\em IEEE Robotics \& Automation Magazine}, 24(3):75--83, 2017.

\bibitem{do2018soft}
Thanh~Nho Do, Hung Phan, Thuc-Quyen Nguyen, and Yon Visell.
\newblock Soft electromagnetic actuators: Miniature soft electromagnetic
  actuators for robotic applications (adv. funct. mater. 18/2018).
\newblock {\em Advanced Functional Materials}, 28(18):1870116, 2018.

\bibitem{follador2015bistable}
M~Follador, AT~Conn, and J~Rossiter.
\newblock Bistable minimum energy structures (bimes) for binary robotics.
\newblock {\em Smart Materials and Structures}, 24(6):065037, 2015.

\bibitem{follador2015design}
Maurizio Follador, Matteo Cianchetti, and Barbara Mazzolai.
\newblock Design of a compact bistable mechanism based on dielectric elastomer
  actuators.
\newblock {\em Meccanica}, 50(11):2741--2749, 2015.

\bibitem{galloway2016soft}
Kevin~C Galloway, Kaitlyn~P Becker, Brennan Phillips, Jordan Kirby, Stephen
  Licht, Dan Tchernov, Robert~J Wood, and David~F Gruber.
\newblock Soft robotic grippers for biological sampling on deep reefs.
\newblock {\em Soft robotics}, 3(1):23--33, 2016.

\bibitem{glick2018soft}
Paul Glick, Srinivasan~A Suresh, Donald Ruffatto, Mark Cutkosky, Michael~T
  Tolley, and Aaron Parness.
\newblock A soft robotic gripper with gecko-inspired adhesive.
\newblock {\em IEEE Robotics and Automation Letters}, 3(2):903--910, 2018.

\bibitem{hawkes2015grasping}
Elliot~W Hawkes, David~L Christensen, Amy~Kyungwon Han, Hao Jiang, and Mark~R
  Cutkosky.
\newblock Grasping without squeezing: Shear adhesion gripper with fibrillar
  thin film.
\newblock In {\em Robotics and Automation (ICRA), 2015 IEEE International
  Conference on}, pages 2305--2312. IEEE, 2015.

\bibitem{ho2017design}
Van Ho and Shinichi Hirai.
\newblock Design and analysis of a soft-fingered hand with contact feedback.
\newblock {\em IEEE Robotics and Automation Letters}, 2(2):491--498, 2017.

\bibitem{homberg2015haptic}
Bianca~S Homberg, Robert~K Katzschmann, Mehmet~R Dogar, and Daniela Rus.
\newblock Haptic identification of objects using a modular soft robotic
  gripper.
\newblock In {\em Intelligent Robots and Systems (IROS), 2015 IEEE/RSJ
  International Conference on}, pages 1698--1705. IEEE, 2015.

\bibitem{kenaley1989electrorheological}
Gary~L Kenaley and Mark~R Cutkosky.
\newblock Electrorheological fluid-based robotic fingers with tactile sensing.
\newblock In {\em Robotics and Automation, 1989. Proceedings., 1989 IEEE
  International Conference on}, pages 132--136. IEEE, 1989.

\bibitem{kim2013soft}
Sangbae Kim, Cecilia Laschi, and Barry Trimmer.
\newblock Soft robotics: a bioinspired evolution in robotics.
\newblock {\em Trends in biotechnology}, 31(5):287--294, 2013.

\bibitem{kim2010towards}
Seung-Won Kim, Je-Sung Koh, Maenghyo Cho, and Kyu-Jin Cho.
\newblock Towards a bio-mimetic flytrap robot based on a snap-through
  mechanism.
\newblock In {\em Biomedical Robotics and Biomechatronics (BioRob), 2010 3rd
  IEEE RAS and EMBS International Conference on}, pages 534--539. IEEE, 2010.

\bibitem{krahn2017soft}
Jeffrey~M Krahn, Francesco Fabbro, and Carlo Menon.
\newblock A soft-touch gripper for grasping delicate objects.
\newblock {\em IEEE/ASME Transactions on Mechatronics}, 22(3):1276--1286, 2017.

\bibitem{lau2017dielectric}
Gih-Keong Lau, Kim-Rui Heng, Anansa~S Ahmed, and Milan Shrestha.
\newblock Dielectric elastomer fingers for versatile grasping and nimble
  pinching.
\newblock {\em Applied Physics Letters}, 110(18):182906, 2017.

\bibitem{liu2018optimal}
Chih-Hsing Liu, Ta-Lun Chen, Chen-Hua Chiu, Mao-Cheng Hsu, Yang Chen, Tzu-Yang
  Pai, Wei-Geng Peng, and Yen-Pin Chiang.
\newblock Optimal design of a soft robotic gripper for grasping unknown
  objects.
\newblock {\em Soft robotics}.

\bibitem{luo2017toward}
Ming Luo, Erik~H Skorina, Weijia Tao, Fuchen Chen, Selim Ozel, Yinan Sun, and
  Cagdas~D Onal.
\newblock Toward modular soft robotics: Proprioceptive curvature sensing and
  sliding-mode control of soft bidirectional bending modules.
\newblock {\em Soft robotics}, 4(2):117--125, 2017.

\bibitem{martinez2014soft}
Ramses~V Martinez, Ana~C Glavan, Christoph Keplinger, Alexis~I Oyetibo, and
  George~M Whitesides.
\newblock Soft actuators and robots that are resistant to mechanical damage.
\newblock {\em Advanced Functional Materials}, 24(20):3003--3010, 2014.

\bibitem{noh2012flea}
Minkyun Noh, Seung-Won Kim, Sungmin An, Je-Sung Koh, and Kyu-Jin Cho.
\newblock Flea-inspired catapult mechanism for miniature jumping robots.
\newblock {\em IEEE Transactions on Robotics}, 28(5):1007--1018, 2012.

\bibitem{pfeifer2009morphological}
Rolf Pfeifer and Gabriel G{\'o}mez.
\newblock Morphological computation--connecting brain, body, and environment.
\newblock In {\em Creating brain-like intelligence}, pages 66--83. Springer,
  2009.

\bibitem{rothemund2018soft}
Philipp Rothemund, Alar Ainla, Lee Belding, Daniel~J Preston, Sarah Kurihara,
  Zhigang Suo, and George~M Whitesides.
\newblock A soft, bistable valve for autonomous control of soft actuators.
\newblock {\em Science Robotics}, 3(16):eaar7986, 2018.

\bibitem{shan2013soft}
Wanliang Shan, Tong Lu, and Carmel Majidi.
\newblock Soft-matter composites with electrically tunable elastic rigidity.
\newblock {\em Smart Materials and Structures}, 22(8):085005, 2013.

\bibitem{shepherd2013soft}
Robert~F Shepherd, Adam~A Stokes, Rui~MD Nunes, and George~M Whitesides.
\newblock Soft machines that are resistant to puncture and that self seal.
\newblock {\em Advanced Materials}, 25(46):6709--6713, 2013.

\bibitem{shih2017custom}
Benjamin Shih, Dylan Drotman, Caleb Christianson, Zhaoyuan Huo, Ruffin White,
  Henrik~I Christensen, and Michael~T Tolley.
\newblock Custom soft robotic gripper sensor skins for haptic object
  visualization.
\newblock In {\em Intelligent Robots and Systems (IROS), 2017 IEEE/RSJ
  International Conference on}, pages 494--501. IEEE, 2017.

\bibitem{shimoga1996soft}
Karun~B. Shimoga and Andrew~A. Goldenberg.
\newblock Soft robotic fingertips: Part i: A comparison of construction
  materials.
\newblock {\em The International Journal of Robotics Research}, 15(4):320--334,
  1996.

\bibitem{shintake2018soft}
Jun Shintake, Vito Cacucciolo, Dario Floreano, and Herbert Shea.
\newblock Soft robotic grippers.
\newblock {\em Advanced Materials}, page 1707035.

\bibitem{shintake2016versatile}
Jun Shintake, Samuel Rosset, Bryan Schubert, Dario Floreano, and Herbert Shea.
\newblock Versatile soft grippers with intrinsic electroadhesion based on
  multifunctional polymer actuators.
\newblock {\em Advanced Materials}, 28(2):231--238, 2016.

\bibitem{shintake2015variable}
Jun Shintake, Bryan Schubert, Samuel Rosset, Herbert Shea, and Dario Floreano.
\newblock Variable stiffness actuator for soft robotics using dielectric
  elastomer and low-melting-point alloy.
\newblock In {\em Intelligent Robots and Systems (IROS), 2015 IEEE/RSJ
  International Conference on}, pages 1097--1102. IEEE, 2015.

\bibitem{song2014soft}
Sukho Song and Metin Sitti.
\newblock Soft grippers using micro-fibrillar adhesives for transfer printing.
\newblock {\em Advanced Materials}, 26(28):4901--4906, 2014.

\bibitem{soter2018bodily}
Gabor Soter, Andrew Conn, Helmut Hauser, and Jonathan Rossiter.
\newblock Bodily aware soft robots: integration of proprioceptive and
  exteroceptive sensors.
\newblock In {\em 2018 IEEE International Conference on Robotics and Automation
  (ICRA)}, pages 2448--2453. IEEE, 2018.

\bibitem{taccola2015toward}
Silvia Taccola, Francesco Greco, Edoardo Sinibaldi, Alessio Mondini, Barbara
  Mazzolai, and Virgilio Mattoli.
\newblock Toward a new generation of electrically controllable hygromorphic
  soft actuators.
\newblock {\em Advanced Materials}, 27(10):1668--1675, 2015.

\bibitem{tavakoli2017autonomous}
Mahmoud Tavakoli, Pedro Lopes, Jo{\~a}o Louren{\c{c}}o, Rui~Pedro Rocha, Luana
  Giliberto, An{\'\i}bal~T de~Almeida, and Carmel Majidi.
\newblock Autonomous selection of closing posture of a robotic hand through
  embodied soft matter capacitive sensors.
\newblock {\em IEEE Sensors Journal}, 17(17):5669--5677.

\bibitem{Thurutheleaav1488}
Thomas~George Thuruthel, Benjamin Shih, Cecilia Laschi, and Michael~Thomas
  Tolley.
\newblock Soft robot perception using embedded soft sensors and recurrent
  neural networks.
\newblock {\em Science Robotics}, 4(26), 2019.

\bibitem{truby2018soft}
Ryan~L Truby, Michael Wehner, Abigail~K Grosskopf, Daniel~M Vogt, Sebastien~GM
  Uzel, Robert~J Wood, and Jennifer~A Lewis.
\newblock Soft somatosensitive actuators via embedded 3d printing.
\newblock {\em Advanced Materials}, 30(15):1706383, 2018.

\bibitem{wang2018toward}
Hongbo Wang, Massimo Totaro, and Lucia Beccai.
\newblock Toward perceptive soft robots: Progress and challenges.
\newblock {\em Advanced Science}, 5(9):1800541, 2018.

\bibitem{wang2016soft}
YuZhe Wang, Ujjaval Gupta, Nachiket Parulekar, and Jian Zhu.
\newblock A soft gripper of fast speed and low energy consumption.
\newblock {\em Science China Technological Sciences}, pages 1--8, 2016.

\bibitem{wang2017prestressed}
Zhongkui Wang, Yuuki Torigoe, and Shinichi Hirai.
\newblock A prestressed soft gripper: design, modeling, fabrication, and tests
  for food handling.
\newblock {\em IEEE Robotics and Automation Letters}, 2(4):1909--1916, 2017.

\bibitem{wingert2006design}
Andreas Wingert, Matthew~D Lichter, and Steven Dubowsky.
\newblock On the design of large degree-of-freedom digital mechatronic devices
  based on bistable dielectric elastomer actuators.
\newblock {\em IEEE/ASME transactions on mechatronics}, 11(4):448--456, 2006.

\bibitem{yap2016high}
Hong~Kai Yap, Hui~Yong Ng, and Chen-Hua Yeow.
\newblock High-force soft printable pneumatics for soft robotic applications.
\newblock {\em Soft Robotics}, 3(3):144--158, 2016.

\bibitem{zhou2015soft}
Xuance Zhou, Carmel Majidi, and Oliver~M O’Reilly.
\newblock Soft hands: an analysis of some gripping mechanisms in soft robot
  design.
\newblock {\em International Journal of Solids and Structures}, 64:155--165,
  2015.

\end{thebibliography}
\end{document}